\documentclass[conference]{IEEEtran}
\usepackage{graphicx}%------------------------------added by ZY
\usepackage{subfigure}
\usepackage{float}%---------------------------------added by ZY
\usepackage{mathrsfs}%------------------------------added by ZY
\usepackage{amsfonts}
\usepackage{array}
\usepackage{amssymb,amsmath}
\usepackage{longtable}
\usepackage{rotating}
\usepackage{multirow}
\usepackage{epstopdf}
\usepackage[lined,boxed,commentsnumbered, ruled]{algorithm2e}
\usepackage{cite}
\usepackage{float}
\usepackage{stfloats}
\usepackage[colorlinks,
            linkcolor=black,
            anchorcolor=black,
            urlcolor=blue,
            citecolor=black]{hyperref}
\usepackage[hyphenbreaks]{breakurl}
\usepackage{geometry}
\renewcommand{\baselinestretch}{1}
\def\BibTeX{{\rm B\kern-.05em{\sc i\kern-.025em b}\kern-.08em
    T\kern-.1667em\lower.7ex\hbox{E}\kern-.125emX}}

\renewcommand{\baselinestretch}{1}
\newcommand{\methodname}{{{DActAHM}}}    
\begin{document}

%\title{Self-Adaptive Health Monitoring in Urban Smart Ecosystems: A Deep Reinforcement Learning Approach}
\title{Deep Reinforcement Learning Empowered Activity-Aware Dynamic Health Monitoring Systems}
\author{\IEEEauthorblockN{Ziqiaing Ye\textsuperscript{1}, Yulan Gao\textsuperscript{2}, Yue Xiao\textsuperscript{1}, 
Zehui Xiong\textsuperscript{3}, and Dusit Niyato\textsuperscript{2} }
\IEEEauthorblockA{\textsuperscript{1}{National Key Laboratory of Wireless Communications} \\
{University of Electronic Science and Technology of China, Chengdu, 611731, China}\\
\IEEEauthorblockA{\textsuperscript{2}{School of Computer Science and Engineering,} {Nanyang Technological University, 639798, Singapore}}
\IEEEauthorblockA{\textsuperscript{3}{ Information Systems Technology and Design Pillar,} {Singapore University of Technology and Design, 487372, Singapore}}
{Email: xiaoyue@uestc.edu.cn}}
}

\maketitle

\begin{abstract} 
In smart healthcare, health monitoring utilizes diverse tools and technologies to analyze patients' real-time biosignal data, enabling immediate actions and interventions.
%In the evolving domain of smart healthcare, the proliferation of wearable devices has revolutionized real-time health monitoring.
Existing monitoring approaches were designed on the premise that medical devices track several health metrics concurrently, tailored to their designated functional scope. This means that they report all relevant health values within that scope, which can result in excess resource use and the gathering of extraneous data due to monitoring irrelevant health metrics. 
%However, the prevailing uniform monitoring approach often culminates in unnecessary resource consumption and accumulation of superfluous data.
In this context, we propose \underline{D}ynamic \underline{Act}ivity-\underline{A}ware \underline{H}ealth \underline{M}onitoring strategy (\methodname{}) for striking a balance between optimal monitoring performance and cost efficiency, a novel framework based on Deep Reinforcement Learning (DRL) and SlowFast Model to ensure precise monitoring based on users' activities.
%dovetails monitoring with users' real-time activities. 
Specifically, with the SlowFast Model, \methodname{} efficiently identifies individual activities and captures these results for enhanced processing. 
%Specifically, by leveraging the SlowFast model, we proficiently detect individuals' activities, subsequently storing these results for enhanced processing.
Subsequently, \methodname{} refines health metric monitoring in response to the identified activity by incorporating a DRL framework. 
%Through the integration of a deep reinforcement learning framework, \methodname{} meticulously adjusts health metric monitoring in tandem with the discerned activity.
Extensive experiments comparing \methodname{} against three state-of-the-art approaches demonstrate it achieves 27.3\% higher gain than the best-performing baseline that fixes monitoring actions over timeline. 
%Rigorous simulations elucidate the superior efficiency and potency of our strategy, manifesting noteworthy enhancements in contrast to conventional methods.
%By adeptly amalgamating activity recognition with pertinent health metrics, our proposition heralds a new era of resourceful, context-sensitive health monitoring in contemporary urban settings.

\end{abstract}


\section{Introduction}
In the rapidly evolving field of smart healthcare,  technological advancements are reshaping medical practices and patient care \cite{ghazal2021iot}, with wearable smart devices at the forefront of this paradigm shift.
Yet, with the widespread adoption of wearables producing an ever-growing volume of data, the main challenge in smart healthcare is meeting the soaring demands of communication and computational resources to maintain optimal Quality of Service (QoS).  
Rising to this challenge means that the capacity of healthcare system must greatly expand to process  the vast data, but all within an extremely tight energy budget.
A promising solution to this dilemma is to optimize health monitoring \cite{mora2017iot}, which employs a diverse array of tools and technologies.  
Health monitoring, integral to smart healthcare, empowers preventive care, fosters data-driven decisions, enhances patient engagement, and offers cost-effective, personalized healthcare solutions while reducing strain on traditional healthcare infrastructures \cite{wu2021internet}.
%The essence of health monitoring revolves around its foundational principle: uninterrupted, real-time tracking of essential health metrics, facilitating swift responses and timely interventions. 
%As we delve deeper into its significance, the imperative of optimizing health monitoring for both efficacy and efficiency becomes undeniably clear\cite{talpur2015energy}.

The essence of health monitoring revolves around its foundational principle: uninterrupted, real-time tracking of essential health metrics, facilitating swift responses and timely interventions.
Specifically, health monitoring systems gather and analyze data to understand an individual's health. This data aids in detecting diseases early, recommending personalized treatments, and helping patients manage their health. As more data is collected, it reveals health trends, enabling proactive care and predictive insights. 
%In essence, data drives health monitoring, and the insights derived from this monitoring enhance the methods of data collection and interpretation.
Thus, the relationship between health monitoring and data is foundational to healthcare. Given the constant growing of data and the requirement for adaptive and optimized healthcare responses, introducing Deep Reinforcement Learning (DRL) methods into monitoring systems can be transformative \cite{verma2022fetch}. 
Among the early contributions in this area, \cite{gupta2019reinforcement} discussed the potential of RL (Reinforcement Learning) -based solutions in Wireless Body Area Network (WBAN).
In \cite{yu2021reinforcement}, some of the RL models that are used in chronic and critical diseases treatments were introduced. 
Further, the authors of \cite{zhu2020basal} developed a DRL model for optimizing single-hormone (insulin) and dual-hormone (insulin and glucagon) delivery.
In \cite{wang2013online}, the Electroencephalogram (EEG) recordings are classified into normal or preseizure baseline patterns using a K-Nearest Neighbor (KNN) classifier.
Then, an RL approach adaptively refines the normal and preseizure baseline patterns based on feedback from prediction outcomes.
%Then, an RL approach is used to adaptively update the normal and preseizure baseline patterns according to the feedback from the prediction result.
Recently, the authors of \cite{guez2008adaptive} 
employed an RL-based approach to dynamically determine an optimal neurostimulation strategy for epilepsy treatment.
Building on these foundations, a lot of literature has concentrated on devising monitoring techniques to meet diverse requirements \cite{kondaka2022intensive,alshamrani2022iot}.
%For example, \cite{hu2019algorithm} proposed a four-element algorithm strategy to integrate physiologic data with clinical data from Electronic Health Record (EHR) to utilize temporal patterns in data streams.
%Traditional methods, typically fixed in their approach, provide regular monitoring procedures\cite{kaur2020internet}, regardless of individual differences.
%On the other hand, advanced techniques, spurred by the digital revolution, leverage intricate algorithms, for example deep learning\cite{verma2022fetch}, to provide more granulated insights, aiming to cater to specific patient profiles and conditions.
%For example, \cite{hu2019algorithm} proposed a four-element algorithm strategy to integrate physiologic data with clinical data from EHR to utilize temporal patterns in data streams.
However, the aforementioned health monitoring approaches were designed under an assumption that  
medical devices track all health metrics simultaneously within their specific functional range.
Moreover, given the vast data output from medical devices (e.g., medical bracelets), monitoring all health metrics indiscriminately and simultaneously can result in significant energy consumption. 
As a result, a reasonable monitoring strategy should essentially exhibit two qualities: selective attention to basic health metrics and dynamic adaptability to adjust these metrics based on the device's performance capabilities.

To bridge these important gaps, we propose Dynamic Activity-Aware Health Monitoring (\methodname{}) strategy. It utilizes SlowFast\footnote{An AI-driven model with full utilization of the capabilities of video-capturing devices \cite{fan2020pyslowfast}.} to recognize and predict individual activities from copious video data streams.  
Subsequently, we design a system that identifies and classifies activities in real-time by leveraging DRL models and merging multi-modal data sources. 
Under our problem setting, once an activity is detected, \methodname{} adaptively adjusts its monitoring to capture pertinent health metrics, minimizing data redundancy and computational overhead. 
The proposed \methodname{} emphasizes the collection of relevant health metrics, facilitating efficient data storage and efficient health analysis. 
We conduct extensive experiments to assess and compare the proposed \methodname{} against three state-of-the-art approaches, then exhibit that \methodname{} outperforms the
best \hyperref[ALG:3]{baseline} by 27.3\% improvement in gain. 
%our study introduces a novel methodology that harnesses the potential of the smart healthcare landscape. We capitalize on the advanced capabilities of video-capturing devices, now augmented with superior sensors and AI-driven algorithms called SlowFast \cite{fan2020pyslowfast}, to meticulously recognize and predict individual activities from copious video data streams. 
%Utilizing cutting-edge deep learning models and merging multi-modal data sources, we design a system that identifies and classifies activities in real-time.
%Upon detection, the system dynamically tailors its monitoring to the relevant health metrics, reducing data redundancy and computational overhead.
%This focused approach ensures meaningful health metric collection, promoting efficient storage and swift data analysis. Our initiative strives for a more responsive, trustworthy, and user-centric experience in smart healthcare.

\section{System Model}
\subsection{System Overview}

\begin{figure*}[!t]
\centering
\includegraphics[width=1\textwidth]{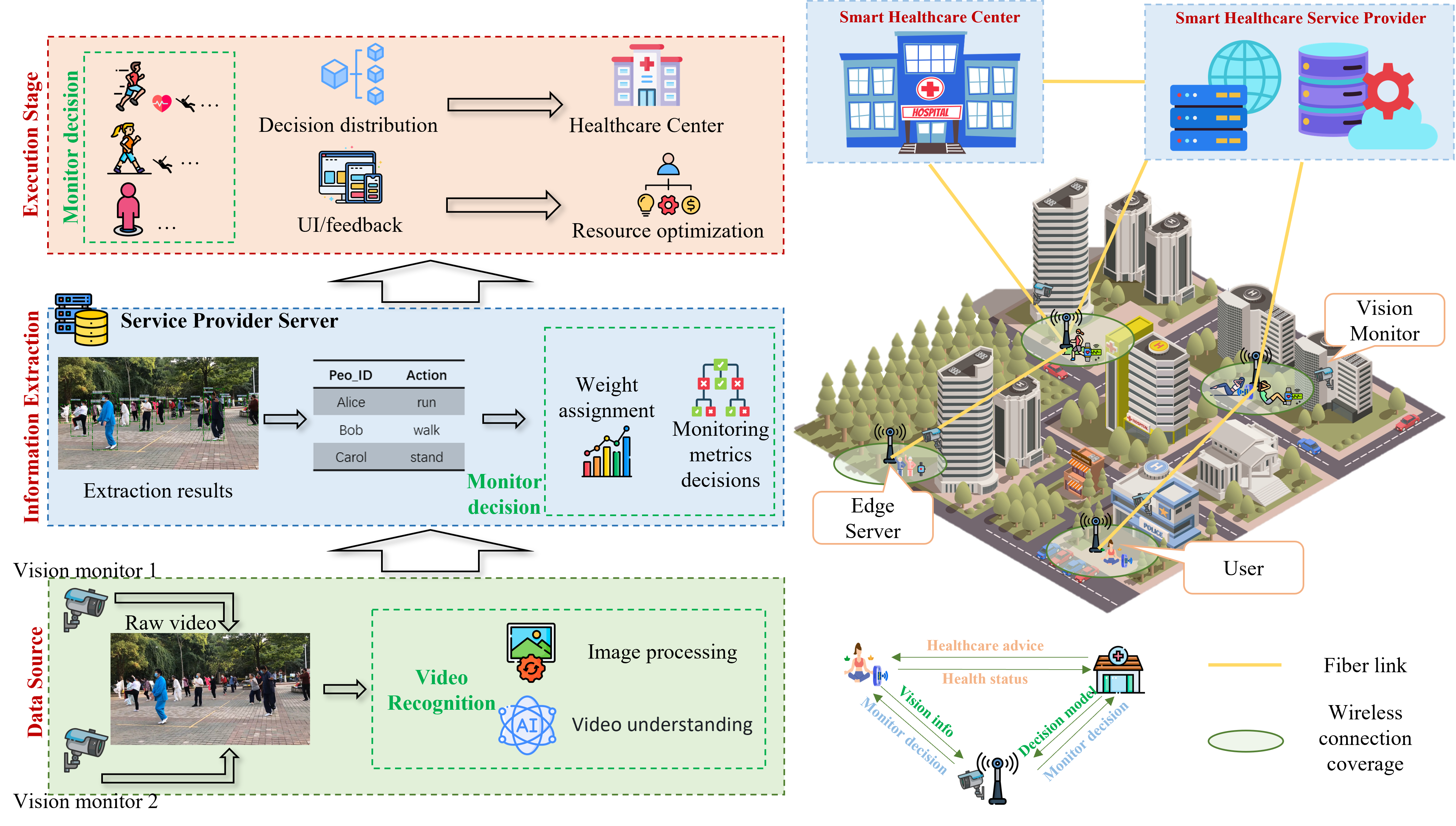}
\caption{A comprehensive overview of a sophisticated smart healthcare system operating in an urban setting.}
\label{fig:1}
\end{figure*}

Consider a smart city framework as depicted in Figure \ref{fig:1},  wherein an extensive network of visual perception devices is employed across all public domains, including but not limited to hospitals, athletic tracks, parks, and various recreational areas. 
The optimal strategy for health monitoring will be realized by recognizing and predicting user activities through these devices.
%In this way, a smart city framework can ensure exhaustive visual coverage.
%we consider a smart city framework encompassing a specific region populated by $N$ individuals, each user wears a smart wearable device.

Consider $N$ individuals located within a specific area of the smart city,  and each user wears a smart wearable device.  
We consider that the same index set of ${\mathcal N}=\{1, 2, \ldots, N\}$ is used by users and their wearable devices. 
Building on \cite{7742297}, we assume that each user is doing an activity, with the entire scope of possible activities are contained within a set ${\mathcal G}=\{1, 2, \ldots, G\}$.     
Let ${\mathbf a}=\{a_n| a_n\in{\mathcal G}\}_{n=1}^N$ denote the activity profile of all users.
The set of health monitoring metrics is denoted by ${\mathcal M}=\{1, 2, \ldots, M\}$.

\subsection{Dynamic Monitoring Strategy Model}
We first outline the dynamic monitoring strategy model within the smart city. 
Building on established monitoring frameworks \cite{swamy2019esmart}, we attempt to leverage visual capture technologies with the aim of enhancing and refining health monitoring strategies.
The whole procedure is systematically delineated in Figure \ref{fig:1}, initiating at the Data Source stage.
Here, massively deployed surveillance cameras capture raw video footage, recording the activities of individuals across the smart city.
After collecting the raw visual data, it proceeds to an intensive Video Recognition phase.
An Information Extraction stage is then triggered at the Provider Server. The particular activities of all users are precisely identified and recorded. 
In the Monitoring Decision phase, the monitoring metric selection is determined according to the identified users' activities. 
Once the monitoring decisions are broadcast from the server during the Execution Stage, the users will receive personalized health monitoring strategies suitable for their current activities, thereby ensuring effective resource allocation and enhanced system performance.

%In the Monitoring Decision phase, we propose an intelligent monitoring metric selection approach $F$ that can dynamically tailor the monitoring strategy, adapting in real-time to the detected activities. Acting as the brain of our system, this function seamlessly maps each detected activity $g$ to the most relevant subset of health metrics $F(g)$. This ensures that the user receives relevant health information in real-time, responding to their current activity.

%In this subsection, we introduce the dynamic monitoring strategy model and propose the concept of an intelligent monitoring metric selection approach in our model to dynamically bridge the gap between activity recognition and health monitoring. This ensures that as users thrive in their daily routines, their wearable devices are in lockstep, providing relevant, real-time health insights. It's a fusion of adaptability with precision, designed for the modern smart city inhabitant.

\subsection{Relevance and Cost of Monitoring Metrics}
Upon introducing the foundation framework of dynamic monitoring strategies, here we focus on developing the relevance and cost of the monitoring metrics.
Given the associations between specific activities and health metrics are commonly considered the cornerstone of health monitoring systems, we will firstly examine the relevance of selected metrics, along with an analysis of the methodologies employed for their quantification.  
For instance, for a jogger, key metrics may encompass heart rate, cadence, and duration of physical exertion. On the other hand, for an individual engaged in meditation practices, parameters such as heart rate variability and respiratory rate emerge as more salient metrics of health and wellness.
This relevance can be referred as the ``demand degree'' or ``value'' that reflects an activity placed on particular health metrics. 
Aided by the introduced demand degree, we could quantitatively determine whether the current monitoring metrics meet the requirements during specific activities.
%Prioritizing monitoring the pivotal metrics enhances the user experience by delivering pertinent data, ensures computational efficiency by conserving resources, and bolsters the accuracy and reliability of the monitoring results.

We deduce the degree of relevance through the utilization of Cosine Similarity \cite{schober2018correlation},  a robust and widely acknowledged metric from the statistical analysis discipline. Let $R(g,F(g))$ denote the degree of relevance for activity $g$ and a sequence of health metrics $F(g)$, where $F(g): {\mathcal G}\mapsto {\mathcal M}$ is a mapping function that assigns each activity $g$ its most relevant subset of health metrics.   
Note that the degree of relevance derived based on Cosine Similarity objectively measures how closely a specific health metric is associated with a specific activity.
By doing so, we can effectively identify the association between activities and health metrics, ensuring that our dynamic monitoring strategy remains efficient and precise.
%By doing so, we can effectively distinguish which health metrics are most relevant for different activities, ensuring our monitoring strategy remains both efficient and pertinent.

As aforementioned that the cost of monitoring metrics is another significant aspect in real-time health monitoring system and arises due to the computational resources required to process and analyze the metrics.
Particularly, in the context of battery-reliant wearable devices, such as smartwatches and fitness trackers, reducing energy consumption becomes crucial  \cite{gao2019dynamic,gao2019dynamicIOT}. 
This is because the overconsumption of energy adversely affects not only the device's life cycle but also its capacity to offer instantaneous feedback to the user, thereby impacting the monitoring performance and user experience. 
Drawing upon \cite{gao2019dynamic,gao2019dynamicIOT}, the computation task for monitoring metric $m\in{\mathcal M}$ at time slot $r$ is denoted by ${\mathcal T}_m=\{D_m[r], c_m[r]\}$, where $D_m[r]$ and $c_m[r]$ represent the datasize of task and the required CPU cycles per datasize to execute the task, respectively.   
%To quantify this computational cost $C(m)$, \cite{gao2023learner} proposed an effective approach based on the device's varying performance and computation task. Following the model in \cite{gao2019dynamic,gao2019dynamicIOT}, we can represent the computation task for each monitoring indicator $m$ at time slot $r$ as a set $\mathcal{T}_m=\{D_m,c_m\}$. This set includes the task's datasize $D_m[r]$ and the required CPU cycles $c_m$ per datasize to execute the task.

We let $f_n$ represent the CPU frequency of device $n$.  
According to \cite{gao2019dynamic,gao2019dynamicIOT}, the energy consumption of device $n$ at time slot $r$ can be determined by
\begin{align}\label{eq:1}
ec_n[r]=\rho f_n^\zeta t_n[r], \forall  n\in\mathcal{N},
\end{align}
where $t_n[r]=\frac{\sum_{m \in F(a_n)}D_m[r] c_m[r]}{f_n}$ and $F(a_n)$ assigns the subset of selected metrics for activity $a_n$ of device $n$. $\rho$ is a constant that depends on the average switched capacitance and the average activity factor, and $\zeta \geq 2$ is a constant.

In smart healthcare, dynamic monitoring strategy design plays a pivotal role in enhancing the effectiveness and efficiency of health monitoring devices. The key objective is to optimize the performance of these devices while ensuring minimal energy consumption and processing delay, both of which are critical parameters in smart healthcare applications. 
To this end, we propose that the computational cost for device $n$ at time slot $r$ can be quantified as a weighted sum of its energy consumption $ec_n[r]$ and the processing delay $t_n[r]$.
\begin{align}\label{eq:2}
    C_n[r]=\mu \cdot ec_n[r]+(1-\mu)t_n[r],
\end{align}
where $0\leq \mu\leq 1$ is a balancing parameter used to calibrate the trade-off between energy consumption and processing delay.
This integrated metric ensures that we holistically capture both the power efficiency and the responsiveness of the device.
By fully considering these two dimensions, we can optimize the device's lifetime and its real-time feedback capabilities, both of which enhance the user experience of smart health monitoring systems.
%By giving due consideration to both these dimensions, we can optimize the device's longevity and its real-time feedback capabilities, both of which are of the essence in ensuring an enhanced user experience in smart health monitoring systems.

\section{Problem Formulation}
In this section, we integrate these fundamental metrics into a structured optimization framework, placing particular emphasis on the formulation of optimization objectives. 
% This will provide us with a clearer view of solutions and ensure the efficiency and accuracy of our health monitoring systems.
  
We first introduce a weight matrix $\pmb\beta \in \mathbb{R}^{N\times M}$ with entries $\beta_{n,m}\in[0, 1]$ representing the assigned weight to health metric $m$ for device $n$.  
It represents the relative importance or priority of each health indicator $m$ given by our monitoring health metrics selection approach $F$. Consequently, $F(a_n)={\pmb\beta}_n$. 
%We introduce a weight matrix $\pmb\beta \in \mathbb{R}^{N\times M}$, designed to represent the relative importance or priority of each health indicator $m$ given by our health metrics monitoring strategy selection approach $F$.
%Each element $\beta_{n,m} \in [0,1]$ in this matrix reflects the assigned weight to a particular health metric $m$ for a specific device $n$.
We then introduce the selection strategy matrix $\pmb\alpha=[{\pmb\alpha}_n]_{n\in{\cal N}} \in \mathbb{R}^{N\times M}$ which can be derived base on $\pmb\beta$.
Specifically, $\pmb\alpha_n \in \mathbb{R}^{1\times M}$ represents the selection strategy vector of device $n$ with entries $\alpha_{n,m}\in\{0, 1\}$ indicating whether device $n$ monitors health metric $m$, which can be expressed as follows: 
\begin{align}\label{eq:3}
    \alpha_{n,m}=\left \{\begin{matrix}
  1,& \text{device }n\text{ is supposed to monitor metric } m;\\
  0,& \text{otherwise.}
\end{matrix}  \right.
\end{align}

The monitoring metrics decisions then become straightforward: if the weight assigned in $\pmb\beta$ for a metric exceeds a predefined threshold that is given by healthcare institution\cite{awotunde2022big}, the corresponding entry in $\pmb\alpha$ is set to $1$. 
This prompts the system to initiate monitoring of the specific metric for device $n$, ensuring a targeted and efficient approach to health data collection.
%signaling the system to monitor that specific indicator for the device $n$. 
Otherwise, it remains $0$, deeming that metric is non-essential for real-time monitoring.
With the introduction of the selection variables $\pmb\beta$ and $\pmb\alpha$, our model gains an additional layer of clarity and precision.
Using the information encapsulated by $\pmb\alpha$, we provide the mathematical expression of degree of relevance, shown as:
%refine our earlier definitions under the assumption that the user of device $n$'s activity is $a_n$.
\begin{align}\label{eq:4}
R(a_n,\pmb \alpha_n[r])&=\frac{I_{a_n} \cdot \pmb \alpha_n[r]}{||I_{a_n}||\cdot||\pmb \alpha_n[r]||}\\ \notag
&=\frac{\sum_{i=1}^MI_{a_n}^i\cdot \alpha_{n,i}}{\sqrt{\sum_{i=1}^M(I_{a_n}^i)^2+\sum_{i=1}^M(\alpha_{n,i})^2}},
\end{align}
where $I_{a_n}$ is the selection matrix of recommended monitoring metrics for activity ${a_n}$ derived from $\mathcal{M}_{a_n}$.
% and $I_{a_n}^i$ refers to the $i$-th value of matrix $I_{a_n}$. 

Furthermore, by incorporating ${\pmb\alpha}$ into our framework, we contextualize our computational costs, 
directly tying them to the specific metrics being monitored. As such, it will provide a more detailed perspective on resource distribution and the patterns of energy consumption. 
%Moreover, we make our computational costs more contextual with the introduction of $\pmb\alpha$, directly tying them to the specific metrics being monitored, thereby offering a more granular insight into the resource allocations and energy consumption patterns.
The time delay in Eq. (\ref{eq:1}) can be rewrite as
\begin{align}\label{eq:5}
    t_n[r]=\frac{\sum \pmb \alpha_n \mathbf D[r] \mathbf c}{f_n},
\end{align}
where $\mathbf D[r] \in \mathbb{R}^{1\times M}$ and $\mathbf c \in \mathbb{R}^{1\times M}$ represent the datasize and the CPU cycles needed per unit datasize for all the monitoring metrics, respectively. 
We also employ the weighted sum method to address the trade-off between monitoring performance and cost using balance parameter $\lambda$.  
Therefore, the objective function can be derived as
\begin{align}\label{eq:6}
    \max_F \sum_{n=1}^N\left(R_n\left(a_n,F\left(a_n\right)\right)-\lambda C_n[r]\right).
\end{align}

\section{\methodname{} Design}
In this section, we leverage RL to solve the optimization problem.
The superiority of RL lies in its natural ability to adaptively learn optimal strategies from interactions with the environment, offering a dynamic and proactive solution that can seamlessly adjust to evolving scenarios, complexities, and uncertainties inherent in real-world health monitoring systems.
Given the continuous nature of the selection strategy in our health monitoring system, i.e., $\pmb\beta \in [0,1]$, Deep Deterministic Policy Gradient (DDPG)\cite{lillicrap2015continuous} emerges as a particularly appropriate solution.
Following the previous study\cite{ye2023cost}, we can define the key factor as follows.

\subsection{State Space}
At the core of our health monitoring system lies the state, encapsulating  the current situational context.  
The state comprises the identified user activity and crucially, the device's CPU frequency $f_n$, which directly impacts its processing capacity.
By accurately recognizing this comprehensive state, our system is empowered to adjust its monitoring mechanisms. 
It also taking into account the users' ongoing activities and devices' computational strength, thereby guaranteeing the delivery of both timely and accurate health insights. 
%ensures that our system can adjust its monitoring mechanisms, factoring in both the user's activity and the device's processing prowess to deliver timely and accurate health insights.
The state space for device $n$ at time slot $r$ is denoted by $\mathcal{S}_n[r]=\{a_n,f_n\}$.

\subsection{Action Space}
The action becomes the system's proactive response to the current state.
As the essence of adaptability, actions are central in ensuring that the monitoring remains relevant, capturing the most pertinent health metrics for the given activity.
The action space involves decisions regarding which health metrics to monitor and its weight for device $n$ at time slot $r$ can be represented as $\mathcal{A}_n[r]=\pmb\beta_n$.

\subsection{Reward}
The reward serves as immediate feedback on the efficacy of the action taken in a particular state.
It could be given based on the objective function given in Eq. \eqref{eq:6}.
For instance, selecting highly relevant metrics for an activity might yield a positive reward, whereas monitoring extraneous or irrelevant metrics might result in a less or even negative reward.
This mechanism guides the learning process, pushing the system towards optimal decisions.
At time slot $r$, the reward can be obtained at device $n$ is denoted by $\mathcal{R}_n[r]=R_n(a_n,F(a_n))-\lambda C_n[r]$.

After that, the detail procedure of \methodname{} is outlined in Algorithm \ref{algorithm:1}.
Initially, the algorithm sets up two models: the SlowFast model for activity detection and a DRL model for adaptive monitoring.
As the primary loop of the algorithm progresses, it captures raw visual data from a vision monitor and employs the SlowFast model to identify the current activity.
If an activity is detected, its result is stored; otherwise, the algorithm moves to its next iteration.
Upon storing an activity, the algorithm feeds the result, along with the device's CPU frequency, into the DRL model, subsequently monitoring the health metrics given by the DRL model.
The aim of this methodology is to ensure health metrics are monitored dynamically, aligning with the detected user activities to maximize utility.
\begin{algorithm}[htpb]
\SetAlgoLined
    \caption{Dynamic Activity-Aware Health Monitoring Strategy (\methodname{})}\label{algorithm:1}
    {\bf Require:} Device's CPU frequency
    
     \KwIn{Initialize SlowFast model for activity detection; DRL model for adaptive monitoring.}
    
    \While {User is detectable}{
     Capture raw visual data from the vision monitor\;
     Use SlowFast model to detect current activity\;
        \eIf {activity detected}
            { Store the activity result\;}
        { Continue to the next iteration\;}
         Send state space to the DRL model to get $\pmb \beta$ to maximize the utility given in Eq. (\ref{eq:6})\;
         Obtain strategy vector $\pmb \alpha$ based on $\pmb \beta$ and predefined threshold\;
         Monitor the metrics based on $\pmb \alpha$\;
}
\KwOut{Adaptively monitored health metrics based on user activities for maximum utility.}
\end{algorithm}

% The complexity of the Algorithm \ref{algorithm:1}, in a simplified manner, would be $\mathcal{O}(T\times(S+D))$ with the assumption that the loop's complexity would be proportional to $T$, SlowFast model's complexity is $\mathcal{O}(S)$ and the DRL's complexity is $\mathcal{O}(D)$.

\section{Simulation Result}
We evaluate the performance of our \methodname{} in the smart healthcare community.
In this paper, we delineate a comprehensive spectrum of human physical activities, categorizing them into a set of $30$ distinct types in the widely used Kinetics-400\cite{kay2017kinetics} dataset.
These activities range from cardio-intensive endeavors like running, jogging, cycling, and jumping rope to strength-based exercises such as weightlifting and squats.
The set also encompasses flexibility and balance exercises like yoga and pilates, leisurely activities including walking and gardening, and even team sports such as football and basketball.

To monitor and understand the health implications of these activities, our system focuses on a set of $10$ pivotal health metrics.
These metrics include the heart rate, which indicates the number of heartbeats per minute, breathing rate reflecting the number of breaths taken, and oxygen saturation which provides insights into the oxygen levels in the bloodstream.

In our DDPG implementation, we the actor and critic networks comprise two fully connected layers. The first layer connects the state input to a hidden layer with 64 neurons, while the second maps these 64 neurons to the action space dimensions.
We use ReLU activation functions for intermediate layers and a sigmoid activation for the output layer to ensure the output action values lie within the $[0,1]$ range.
The discount factor $\gamma$ is set to $0.99$. This ensures that our agent takes into account long-term rewards but does not excessively discount them, striking a balance between immediate and future gains.
A Soft Update Coefficient $\tau=0.005$ is used to ensure a gentle blending of main and target network weights.

Drawing upon \cite{anwer2021evaluation,mukhopadhyay2014wearable}, we assign different monitoring indicator weights to different sports activities to represent the importance of each monitoring indicator in the sports activity.
\begin{figure}[!t]
\centering
\includegraphics[width=0.4\textwidth]{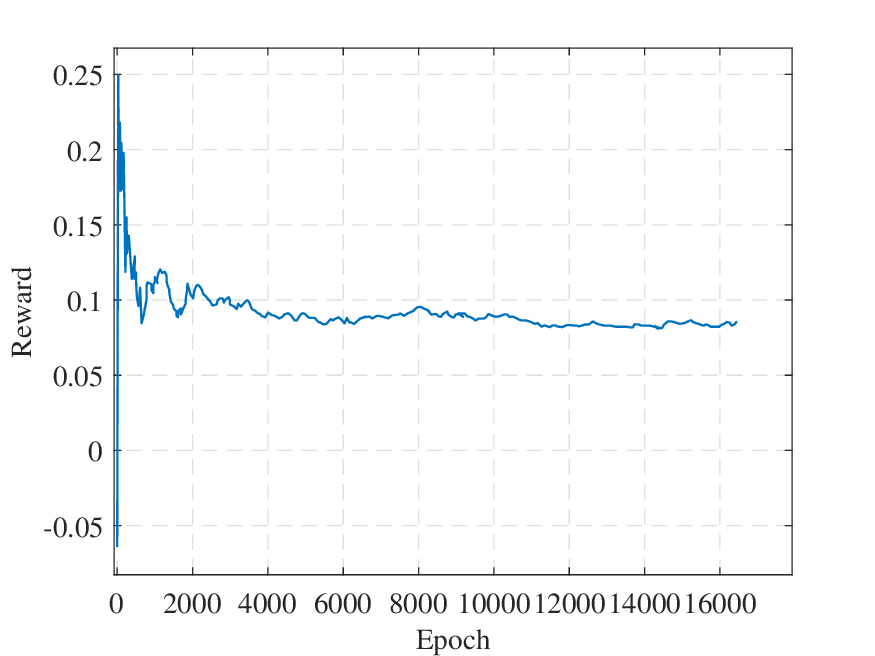}
\caption{DDPG algorithm convergence verification.}
\label{fig:2}
\end{figure}

\begin{figure}[!t]
\centering
\includegraphics[width=0.4\textwidth]{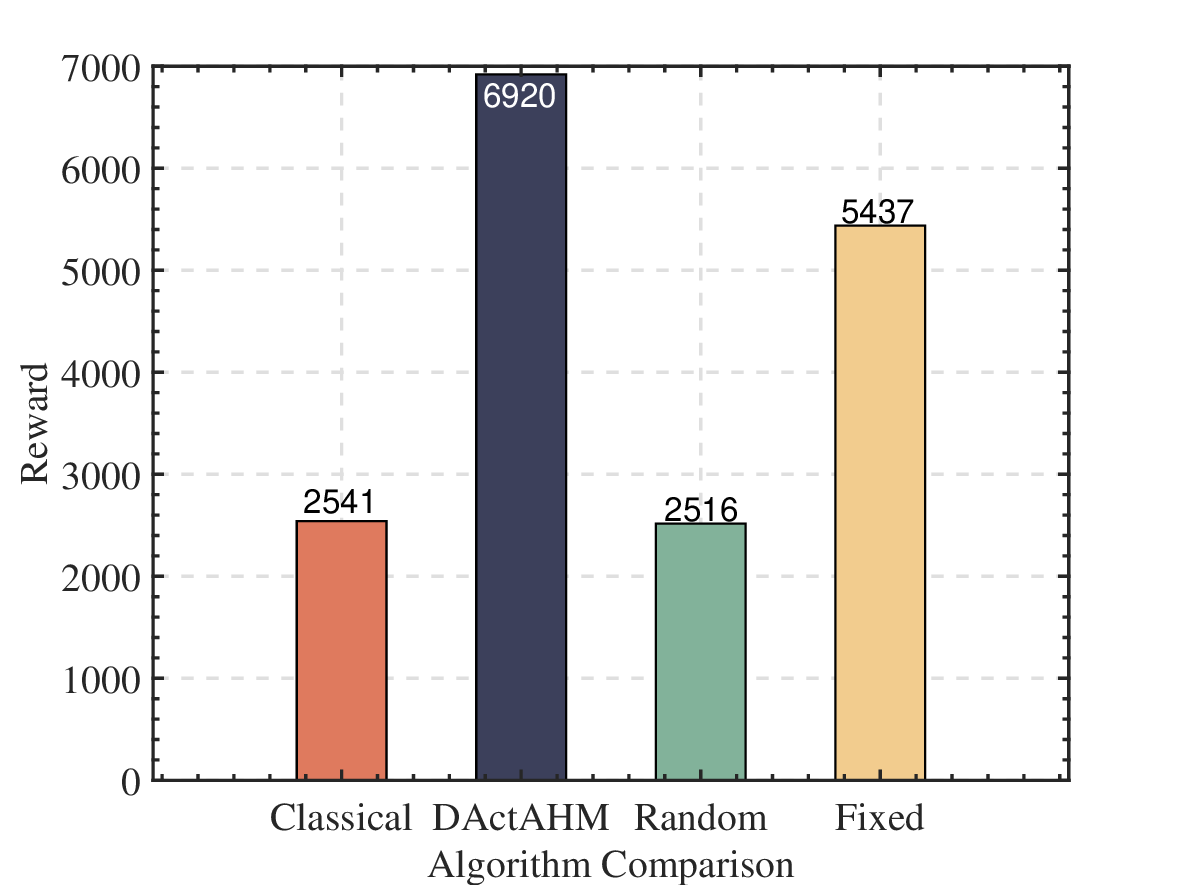}
\caption{Algorithm comparison in static scenario.}
\label{fig:3}
\end{figure}

\begin{figure}[!t]
\centering
\includegraphics[width=0.4\textwidth]{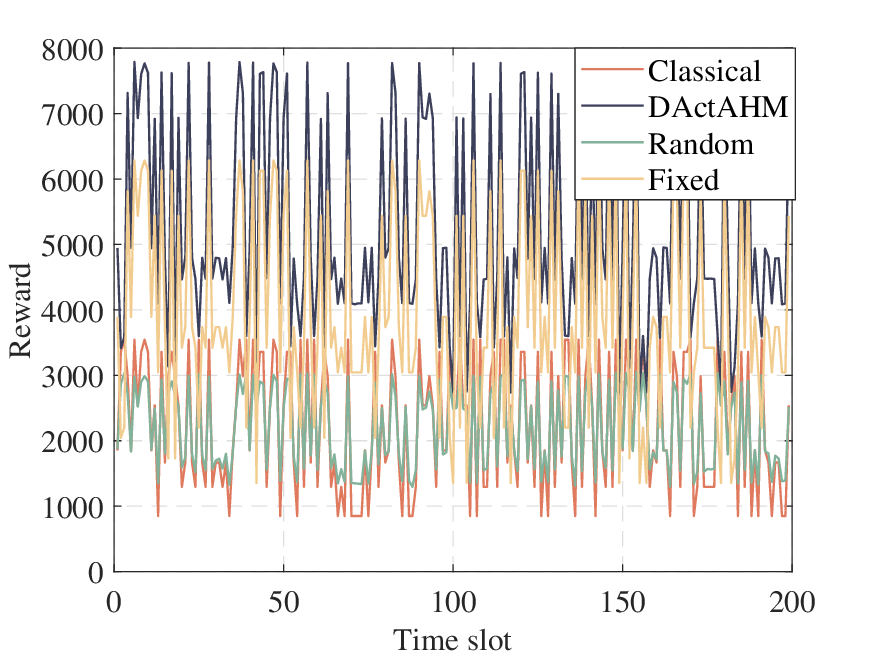}
\caption{Algorithm comparison in dynamic scenario.}
\label{fig:4}
\end{figure}
Figure \ref{fig:2} presents the rewards as the DDPG algorithm iterates through successive epochs.
In the initial stages, there is a swift rise in rewards, indicating rapid early learning.
However, beyond $2000$ epoch, the increase in rewards begins to gradually decreases, eventually approaches to a stable value, indicating that \methodname{} has reached convergence. 
%stabilizing at a constant value, signifying the algorithm's convergence.
In addition to the overarching observation, the early epochs exhibit notable volatility in reward values, typical of the exploration phase in RL.
The rewards, which commence near $0.2$, experience a rapid descent, eventually stabilizing just above $0.05$.
Figure. \ref{fig:2} shows that the DDPG agent's process from rapid initial learning to a phase of consolidation and refinement, culminating in a reliable and positive reward structure.

Figure \ref{fig:3} depicts a comparative evaluation of the rewards garnered through the implementation of the \methodname{} and the following three relevant baselines.
\begin{enumerate}
\item {\bf Classical}\label{ALG:1}: which monitors all metrics.    
\item {\bf Random}\label{ALG:2}: which is characterized by its arbitrary monitoring of all metrics. 
\item {\bf Fixed}\label{ALG:3}: adheres to monitoring a fixed set of metrics. 
\end{enumerate}

As seen in Fig. \ref{fig:3}, the vertical axis represents the reward value, while the horizontal axis enumerates the four strategies. 
It is evident that \methodname{} outperforms the other methods by a significant margin, amassing a reward of $6920$.
In contrast, the Classical approach, which monitors all metrics, yields a substantially lower reward of $2541$.
The reason why the Classic approach has a lower reward is that it monitors all metrics.
Users who are doing non-strenuous exercise will waste too much computing resources because of this monitoring approach.
The Random method, characterized by its arbitrary monitoring of all metrics, results in a reward of 2516, almost mirroring the Classical strategy.
Meanwhile, the Fixed strategy, which adheres to monitoring a fixed set of metrics, achieves a commendable reward of $5437$.
Although the Fixed Strategy inferior to the performance exhibited by \methodname{}, it still exceeds the rewards obtained by Classical and Random methods.
%This value, though inferior to \methodname{}, surpasses both Classical and Random methods.
From the results, we observe the superior efficiency of \methodname{} in the context of monitoring metrics, highlighting its potential for optimized performance in real-world applications.

Figure \ref{fig:4} captures the fluctuation of rewards over time for four distinct strategies.
\methodname{}, consistently hovers at a higher reward throughout the timeline, showcasing its resilience and efficiency in accruing rewards. In many instances, especially towards the later half of the time slots, this strategy appears to outperform the others, signaling its robustness.
In summation, the graph vividly contrasts the dynamism of the four strategies, underscoring the superiority of the \methodname{} in achieving consistently higher rewards across the timeline.

The elevated rewards achieved by \methodname{} can be attributed to its adaptive nature. Unlike the other strategies, \methodname{} optimizes its monitoring strategy dynamically, optimizing based on the immediate context and conditions.
This intelligent adaptation ensures that only the most pertinent metrics are monitored, minimizing resource wastage and, in turn, maximizing rewards.
Comparing the Classical strategy and Fixed strategy offers further insight.
The Classical strategy, which monitors all metrics indiscriminately, may be consuming more resources than necessary.
In scenarios where not all metrics are essential, this method can lead to computational and energy overheads.
The Fixed strategy, while fixed in its monitoring approach, is perhaps focused on a subset of metrics that are generally more relevant across various contexts.
This limited yet contextually apt monitoring could explain why Fixed occasionally outperforms Classical, which, despite its comprehensive monitoring, might be deemed inefficient.

\section{Conclusion}
% In this study, we delved into the evolving world of smart healthcare, exploring the integration of dynamic monitoring strategies with the vast capabilities of wearable devices.
% The primary motivation was to enhance the efficiency of these devices by aligning their monitoring protocols with users' real-time activities.
% Our \methodname{}, hinged on a DDPG model, demonstrated superior adaptability by tailoring the monitoring approach based on the current activity. 
% Comparative analysis revealed that our method consistently outperformed traditional, random, and fixed monitoring strategies.
% The adaptability of the \methodname{} not only conserved computational resources but also amplified the relevance of the health metrics provided to the users.
% The insights gleaned from the simulation results affirm the potential of context-aware, adaptive monitoring in smart healthcare ecosystems.
% As urban environments become increasingly digitized, such innovations will play a pivotal role in ensuring that health monitoring remains both efficient and user-centric.

% In the rapidly advancing domain of smart healthcare, wearable devices have become transformative tools for real-time health monitoring.
% However, many devices have been designed to track a multitude of health metrics, often leading to unnecessary resource consumption and extraneous data accumulation. 
% Addressing this challenge, we introduced the 
The concept of \methodname{} was developed as a novel novel framework to leverage the strengths of DRL and the SlowFast Model, ensuring pinpoint health monitoring attuned to users' specific activities.
Leveraging the SlowFast Model, \methodname{} adeptly recognized user's activities and capitalized on this result for improving monitoring precision.
Furthermore, the integration of a DRL mechanism allows \methodname{} to fine-tune health metrics based on detected activities.
We conducted extensive experiments to assess and compare the proposed \methodname{} against three  methods. Our findings indicated that \methodname{} outperforms the best baseline by 27.3\%  improvement in gain.  
This adaptive capacity not only optimizes computational resources but also boosts the relevancy of the data presented to users.
Such results underscore the immense promise of activity-tailored, adaptive monitoring in the smart healthcare realm, which is poised to play an increasingly integral role in digitally-enhanced urban settings.
\bibliographystyle{ieeetr}
\bibliography{conference_101719}

\end{document}